\pdfoutput=1

\documentclass[11pt]{article}

\usepackage[]{acl}

\usepackage{times}
\usepackage{latexsym}
\usepackage{hyperref}       
\usepackage{url}            
\usepackage{booktabs}       
\usepackage{amsfonts}       
\usepackage{nicefrac}       
\usepackage{microtype}      
\usepackage{xcolor}         
\usepackage{soul}
\usepackage{graphicx}
\usepackage{subcaption}
\usepackage{caption}
\usepackage{bbding}
\usepackage{tikz}
\usepackage{pgf-pie}
\usepackage{bbold}
\usepackage{enumitem}

\usepackage[T1]{fontenc}

\usepackage[utf8]{inputenc}

\usepackage{microtype}

\newcommand{\dataset}{{\bbfamily QAConv}}
\newcommand{\questionNumber}{34,608}
\newcommand{\dialogNumber}{10,259}
\newcommand{\chunkNumber}{18,728}

\title{{\bbfamily QAConv}: Question Answering on Informative Conversations}

\author{%
  Chien-Sheng Wu$^1$, Andrea Madotto$^2$, Wenhao Liu$^1$, Pascale Fung$^2$, Caiming Xiong$^1$\\
  $^1$Salesforce AI Research\\
  $^2$The Hong Kong University of Science and Technology\\
  \texttt{\{wu.jason, wenhao.liu, cxiong\}@salesforce.com} \\ \texttt{amadotto@connect.ust.hk, pascale@ece.ust.hk}
}

\begin{document}
\maketitle
\begin{abstract}
This paper introduces {\dataset},
\footnote{Data and code are available at \url{https://github.com/salesforce/QAConv}}, 
a new question answering (QA) dataset that uses conversations as a knowledge source. We focus on informative conversations, including business emails, panel discussions, and work channels. Unlike open-domain and task-oriented dialogues, these conversations are usually long, complex, asynchronous, and involve strong domain knowledge. 
In total, we collect {\questionNumber} QA pairs from {\dialogNumber} selected conversations with both human-written and machine-generated questions.
We use a question generator and a dialogue summarizer as auxiliary tools to collect and recommend questions.
The dataset has two testing scenarios: chunk mode and full mode, depending on whether the grounded partial conversation is provided or retrieved.
Experimental results show that state-of-the-art pretrained QA systems have limited zero-shot performance and tend to predict our questions as unanswerable. 
Our dataset provides a new training and evaluation testbed to facilitate QA on conversations research.
\end{abstract}

\section{Introduction}
\begin{figure*}[t]
  \centering
  \includegraphics[width=0.8\linewidth]{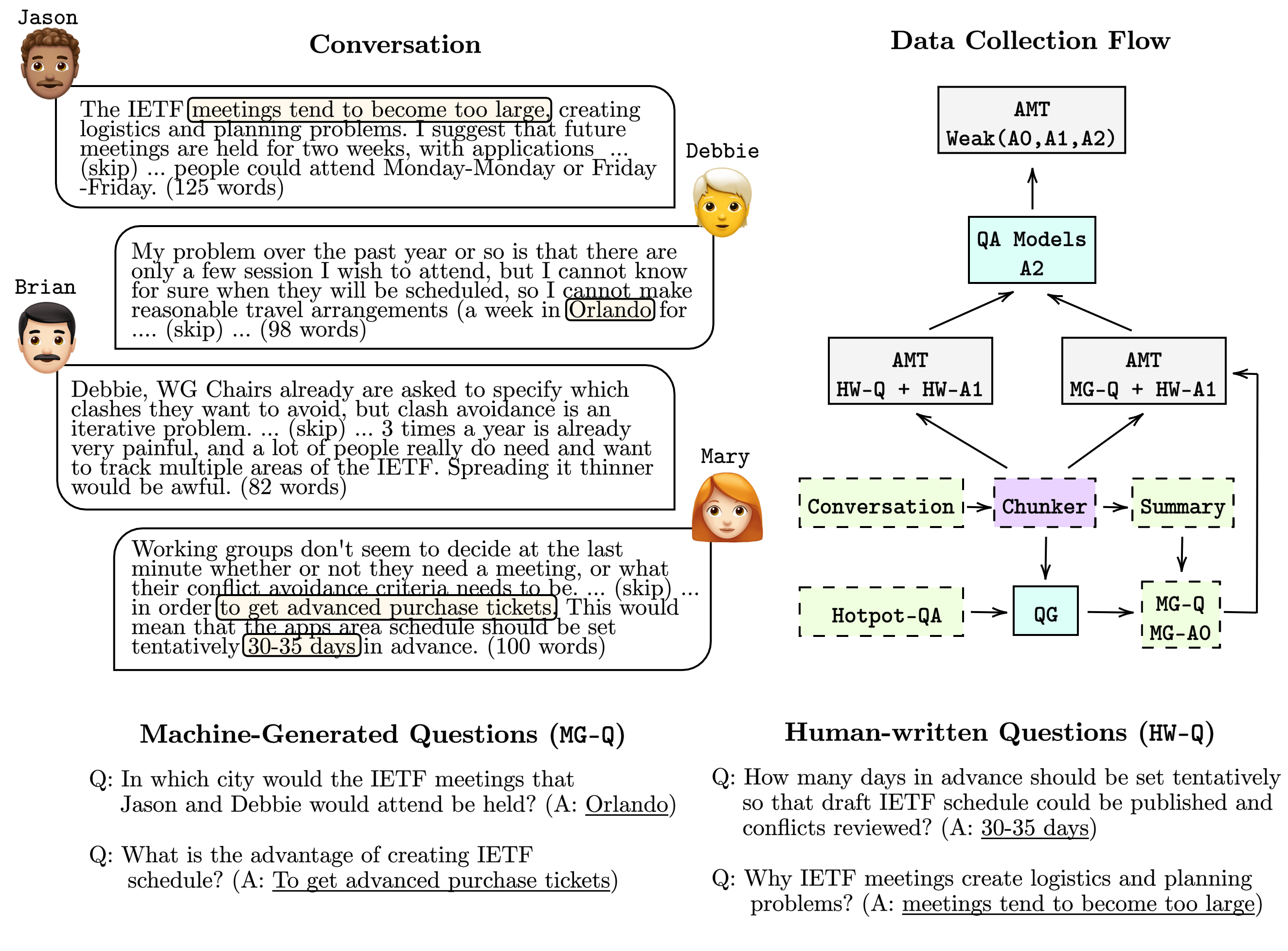}
  \caption{An example of question answering on conversations and the data collection flow.}
  \label{FIG:example}
\end{figure*}

Having conversations is one of the most common ways to share knowledge and exchange information. 
Recently, many communication tools and platforms are heavily used with the increasing volume of remote working, and how to effectively retrieve information and answer questions based on past conversations becomes more and more important. 
In this paper, we focus on QA on conversations such as business emails (e.g., Gmail), panel discussions (e.g., Zoom), and work channels (e.g., Slack). Different from daily chit-chat~\citep{li-etal-2017-dailydialog} and task-oriented dialogues~\citep{budzianowski-etal-2018-multiwoz}, these conversations are usually long, complex, asynchronous, multi-party, and involve strong domain knowledge. We refer to them as informative conversations and an example is shown in Figure~\ref{FIG:example}. 

However, QA research mainly focuses on document understanding (e.g., Wikipedia) not dialogue understanding, and dialogues have significant differences with documents in terms of data format and wording style, and important information is scattered in multiple speakers and turns~\citep{wolf2019transfertransfo,wu-etal-2020-tod}. 
Moreover, existing work related to QA and conversational AI focuses on conversational QA~\citep{reddy-etal-2019-coqa, choi-etal-2018-quac} instead of QA on conversations. Conversational QA has sequential dialogue-like QA pairs that are grounded on a short document paragraph, but what we are more interested in is to have QA pairs grounded on conversations, treating past dialogues as a knowledge source.

QA on conversation has several unique challenges: 1) information is distributed across multiple speakers and scattered among dialogue turns; 2) Harder coreference resolution problem of speakers and entities, and 3) missing supervision as no training data in such format is available.
The most related work to ours is the FriendsQA dataset \citep{yang-choi-2019-friendsqa} and the Molweni dataset~\citep{li2020molweni}. However, the former is built on chit-chat transcripts of TV shows with only one thousand dialogues, and the latter has short conversations in a specific domain (i.e., Ubuntu). The dataset comparison is shown in Table~\ref{Tab:dataset_comparison}.

Therefore, we introduce {\dataset} dataset, sampling {\dialogNumber} conversations from email, panel, and channel data. The longest dialogue sample in our data has 19,917 words (or 32 speakers), coming from a long panel discussion. We segment long conversations into shorter conversational chunks to collect human-written (HW) QA pairs or to modify machine-generated (MG) QA pairs from Amazon Mechanical Turk (AMT). We train a multi-hop question generator and a dialogue summarizer to generate QA pairs. We use QA models to identify uncertain samples and conduct an additional human verification stage. The data collection flow is shown in Figure~\ref{FIG:example}. In total, we collect {\questionNumber} QA pairs.


We construct two testing scenarios: 1) In the chunk mode, a conversational chunk is provided to answer questions, similar to the SQuAD dataset~\citep{rajpurkar-etal-2016-squad}; 2) In the full mode, a conversational-retrieval stage is required before answering questions, similar to the open-domain QA dataset~\citep{chen-yih-2020-openqa}.
We explore several state-of-the-art QA models such as the span extraction RoBERTa-Large model~\citep{liu2019roberta} trained on SQuAD 2.0 dataset, and the generative UnifiedQA model~\citep{khashabi-etal-2020-unifiedqa} trained on eight different QA datasets and showed its generalization ability to 12 unseen QA corpora. 
We investigate the statistic-based BM25~\citep{Robertson1994OkapiAT} retriever and the neural-based dense passage retriever~\citep{karpukhin-etal-2020-dense} trained on Wikipedia (DPR-wiki).
We show zero-shot and finetuning performances in both modes and conduct improvement study and error analysis.


The main contributions of our paper are threefold: 1) {\dataset} provides a new testbed for QA on informative conversations including emails, panel discussions, and work channels. We show the potential of treating long conversations as a knowledge source, and point out a performance gap between QA on documents and QA on conversations; 2) We incorporate question generation (QG) model into the QA data collection, and we show the effectiveness of such approach in human evaluation. 3) We introduce chunk mode and full mode settings for QA on conversations, and our training data enables existing QA models to perform better on dialogue understanding.

\section{{\dataset} Dataset}
Our dataset is collected in four stages: 1) selecting and segmenting informative conversations, 2) generating question candidates by QG models, 3) crowdsourcing question-answer pairs on those conversations/questions, and 4) conducting quality verification and data splits.

\begin{table*}[t]
\centering
\resizebox{0.85\linewidth}{!}{
\begin{tabular}{lccccc}
\hline
\textbf{} & \multicolumn{2}{c}{ {\dataset}} & \textbf{Molweni} & \textbf{DREAM} & \textbf{FriendsQA} \\ \cline{2-3}
 & Full & Chunk &  &  \\ \hline
Source & \multicolumn{2}{c}{Email, Panel, Channel} & Channel & Chit-chat & Chit-chat \\
Domain & \multicolumn{2}{c}{General} & Ubuntu & Daily & TV show \\
Formulation & \multicolumn{2}{c}{Span/Unanswerable} & Span/Unanswerable & Multiple choice & Span \\
Questions & \multicolumn{2}{c}{\textbf{\questionNumber}} & {30,066} & 10,197 & 10,610 \\
Dialogues & {\dialogNumber} & \textbf{\chunkNumber} & 9,754 & 6,444 & 1,222 \\
Avg/Max Words & \textbf{568.8} / \textbf{19,917} & 303.5 / 6,787 & 104.4 / 208 & 75.5 / 1,221 & 277.0 / 2,438 \\
Avg/Max Speakers & 2.8 / \textbf{32} & 2.9 / 14 & \textbf{3.5} / 9 & 2.0 / 2 & 3.9 / 15 \\ \hline
\end{tabular}
}
\caption{Dataset comparison with existing datasets.}
\label{Tab:dataset_comparison}
\end{table*}

\begin{table*}[t]
\centering
\resizebox{\linewidth}{!}{
\begin{tabular}{l|cc|cc|cc}
\hline
\textbf{} & \multicolumn{2}{c|}{\textbf{BC3}} & \multicolumn{2}{c|}{\textbf{Enron}} & \multicolumn{2}{c}{\textbf{Court}} \\
 & Full & Chunk & Full & Chunk & Full & Chunk \\ \hline
Questions & \multicolumn{2}{c|}{174} & \multicolumn{2}{c|}{8,647} & \multicolumn{2}{c}{10,037} \\
Dialogues & 40 & 84 & 3,257 & 4,220 & 125 & 4,923 \\
Avg/Max Words & 514.9 / 1,236 & 245.2 / 593 & 383.6 / 69,13 & 285.8 / 6,787 & 13,143.4 / 19,917 & 330.7 / 1,551 \\
Avg/Max Speakers & 4.8 / 8 & 2.7 / 6 & 2.7 / 10 & 2.2 / 8 & 10.3 / 14 & 2.7 / 7 \\ \hline
\end{tabular}
}
\resizebox{0.6\linewidth}{!}{
\begin{tabular}{l|cc|cc}
\hline
\textbf{} & \multicolumn{2}{c|}{\textbf{Media}} & \multicolumn{2}{c}{\textbf{Slack}} \\
 & Full & Chunk & Full & Chunk \\ \hline
Questions & \multicolumn{2}{c|}{9,753} & \multicolumn{2}{c}{5,997} \\
Dialogues & 699 & 4,812 & 6,138 & 4,689 \\
Avg/Max Words & 2,009.6 / 11,851 & 288.7 / 537 & 247.2 / 4,777 & 307.2 / 694 \\
Avg/Max Speakers & 4.4/ 32 & 2.4 / 11 & 2.5 / 15 & 4.3 / 14 \\ \hline
\end{tabular}
}
\caption{Dataset statistics of different dialogue sources.}
\label{Tab:dataset_statistics}
\end{table*}

\subsection{Data Collection}

\subsubsection{Selection and Segmentation} 
Full data statistics are shown in Table~\ref{Tab:dataset_statistics}. First, we use the British Columbia conversation corpora (BC3)~\citep{JanAAAI08-bc3} and the Enron Corpus~\citep{klimt2004enron} to represent business email use cases. The BC3 is a subset of the World Wide Web Consortium's (W3C) sites that are less technical. We sample threaded Enron emails from~\cite{agarwal-etal-2012-comprehensive}, which were collected from the Enron Corporation.
Second, we select the Court corpus~\citep{danescu2012echoes-court} and the Media dataset~\citep{zhu2021mediasum} as panel discussion data. The Court data is the transcripts of oral arguments before the United States Supreme Court. The Media data is the interview transcriptions from National Public Radio and Cable News Network.
Third, we choose the Slack chats~\citep{openslack} to represent work channel conversations. The Slack data was crawled from several public software-related development channels such as \textit{pythondev\#help}.
All data we use is publicly available and their license and privacy (Section~\ref{sec:license}) information are shown in the Appendix. 

One of the main challenges in our dataset collection is the length of input conversations and thus resulting in very inefficient for crowd workers to work on. For example, on average there are 13,143 words per dialogue in the Court dataset, and there is no clear boundary annotation in a long conversation of a Slack channel. Therefore, we segment long dialogues into short chunks by a turn-based buffer to assure that the maximum number of tokens in each chunk is lower than a fixed threshold, i.e., 512. For the Slack channels, we use the disentanglement script from~\cite{openslack} to split channel messages into separated conversational threads, then we either segment long threads or combine short threads to obtain the final conversational chunks.

\subsubsection{Question Generation}
Synthetic dataset construction has been shown to improve robustness~\cite{gupta2021synthesizing} and improve the complexity of test sets~\cite{feng2021survey}.
We leverage a question generator and a dialogue summarizer to generate and recommend some questions to workers.
We train a T5-Base~\citep{raffel2019exploring} model on HotpotQA~\citep{yang-etal-2018-hotpotqa}, which is a QA dataset featuring natural and multi-hop questions, to generate questions for our conversational chunks. By the second hypothesis, we first train a BART~\citep{lewis-etal-2020-bart} summarizer on News~\citep{narayan-etal-2018-dont-xsum} and dialogue summarization corpora~\citep{gliwa2019samsum} and run QG models on top of the generated summaries. 
\begin{figure*}[t]
    \centering
    \includegraphics[width=0.9\linewidth]{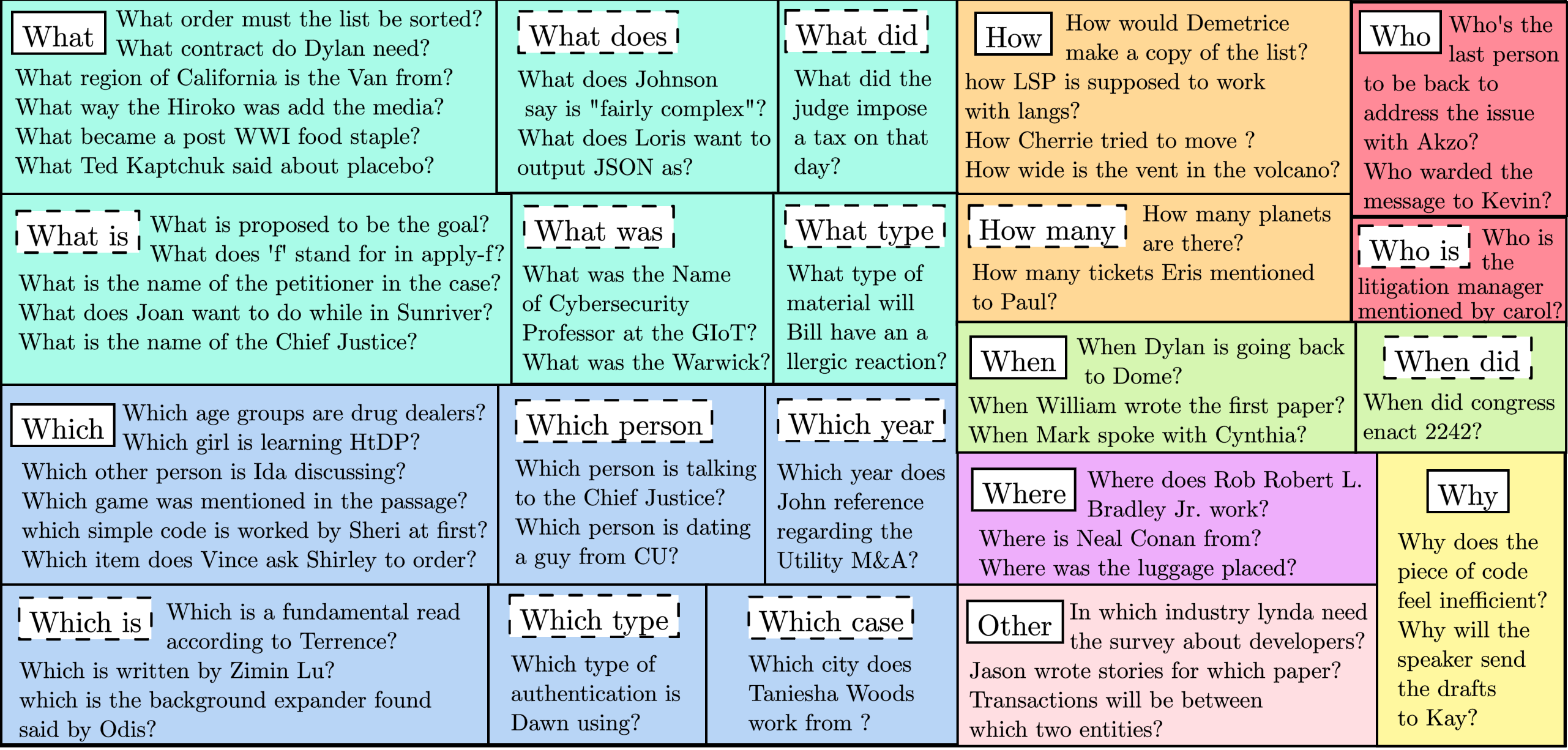}
    \caption{Question type tree map and examples (Best view in color).}
    \label{fig:question_treemap}
\end{figure*}

\begin{table*}[t]
    
    \centering
    \resizebox{\linewidth}{!}{
    \begin{tabular}{ccccccc}
    \hline
     \textbf{QAConv}         & \textbf{Squad 2.0}      & \textbf{QuAC}          & \textbf{CoQA}           & \textbf{Molweni} & \textbf{FriendQA}       & \textbf{DREAM}             \\
    \hline
     what (29.09\%)  & what (49.07\%)  & what (35.67\%) & what (31.02\%)  & what (65.9\%) & what (19.97\%)  & what (53.33\%)     \\
     which (27.21\%) & how (9.54\%)    & did (19.19\%)  & who (13.43\%)   & how (11.4\%) & who (18.1\%)    & how (11.32\%)      \\
     how (11.54\%)   & who (8.36\%)    & how (8.13\%)   & how (9.38\%)    & who (7.54\%) & where (16.07\%) & where (10.29\%)    \\
     who (9.99\%)    & when (6.2\%)    & was (6.05\%)   & did (8.0\%)     & why (5.57\%)& why (15.99\%)   & why (7.94\%)       \\
     when (6.03\%)   & in (4.35\%)     & are (5.45\%)   & where (6.41\%)  & where (5.54\%) & how (15.14\%)   & when (5.05\%)      \\
     where (4.48\%)  & where (3.62\%)  & when (5.43\%)  & was (4.53\%)    & when (1.84\%) & when (11.76\%)  & who (2.89\%)       \\
     why (2.75\%)    & which (2.83\%)  & who (4.62\%)   & when (3.29\%)   & which (1.53\%)& which (0.51\%)  & which (2.84\%)     \\
     in (1.79\%)     & the (2.47\%)    & why (3.11\%)   & why (2.73\%)    & whose (0.12\%) & at (0.34\%)     & the (1.57\%)       \\
     the (1.46\%)    & why (1.58\%)    & where (3.06\%) & is (2.69\%)     & is (0.09\%) & monica (0.34\%) & according (0.59\%) \\
     on (0.38\%)     & along (0.36\%)  & is (1.74\%)    & does (2.09\%)   & did (0.08\%)& whom (0.25\%)   & in (0.49\%)        \\
     \hline
     Other (5.27\%)  & Other (11.62\%) & Other (7.55\%) & Other (16.41\%) & others (0.42\%) & Other (1.52\%)  & Other (3.68\%)     \\
    \hline
    \end{tabular}
    }
    \caption{Question type distributions: Top 10.}
    \label{tab:question_type_top10}
\end{table*}

We filter out generated questions that a QA model can predict the same answers we used in our QG model, which we hypothesize that these questions could be easy questions that we would like to avoid.
Note that our QG model has grounded answers since it is trained to generate questions by giving a text context and an extracted entity.
We hypothesize that these questions are trivial questions in which answers can be easily found, and thus not interesting for our dataset. Examples of our generated multi-hop questions are shown in the Appendix (Table~\ref{Tab:multi-hop-questions}). 

\subsubsection{Crowdsourcing QA Pairs}
We use two strategies to collect QA pairs, human writer and machine generator. We first ask crowd workers to read partial conversations, and then we randomly assign two settings: 1) writing QA pairs themselves or 2) selecting one recommended machine-generated question to answer. We apply several on-the-fly constraints to control the quality of the collected QA pairs: 1) questions should have more than 6 words with a question mark in the end; 2) questions and answers cannot contain first-person and second-person pronouns (e.g., I, you, etc.); 3) answers have to be less than 20 words , and 4) all words have to appear in source conversations. 

We randomly select four MG questions from our question pool and ask crowd workers to answer one of them, without providing any potential answers. They are allowed to modify questions if necessary. To collect unanswerable questions, we ask crowd workers to write questions with at least three entities mentioned in the given conversations but they are not answerable. We pay crowd workers roughly \$8-10 per hour, and the average time to read and write one QA pair is approximately 4 minutes. 

\subsubsection{Quality Verification and Data Splits} 
We design a filter mechanism based on different potential answers: human writer's answers, answer from existing QA models, and QG answers. If all the answers have a pairwise fuzzy matching ratio (FZ-R) scores~\footnote{\url{https://pypi.org/project/fuzzywuzzy}} lower than 75\%, we then run another crowdsourcing round and ask crowd workers to select one of the following options: A) the QA pair looks good, B) the question is not answerable, C) the question has a wrong answer, and D) the question has a right answer but I prefer another answer. We run this step on around 40\% samples which are uncertain. We filter the questions of the (C) option and add answers of the (D) option into the ground truth. In questions marked with option (B), we combine them with the unanswerable questions that we have collected. In addition, we include 1\% random questions (questions that are sampled from other conversations) to the same batch of data collection as a qualification test. We filter crowd workers' results if they fail to indicate such a question as an option (B). Finally, we split the data into 27,287 training samples, 3,660 validation samples, and 3,661 testing samples. There are 4.7\%, 5.1\%, 4.8\% unanswerable questions in train, validation, and test split, respectively.

\subsection{QA Analysis}
In this section, we analyze our collected questions and answers. We first investigate question type distribution and we compare human-written questions and machine-generated questions. We then analyze answers by an existing named-entity recognition (NER) model and a constituent parser.

\subsubsection{Question Analysis} 

\noindent\textbf{Question Type.}
We show the question type tree map in Figure~\ref{fig:question_treemap} and the detailed comparison with other datasets in Table~\ref{tab:question_type_top10}. In {\dataset}, the top 5 question types are what-question (29\%), which-question (27\%), how-question (12\%), who-question (10\%), and when-question (6\%). Comparing to SQuAD 2.0 (49\% what-question), our dataset have a more balanced question distribution. The question distribution of unanswerable questions is different from the overall distribution. The top 5 unanswerable question types are what-question (45\%), why-question (15\%), how-question (12\%), which-question (10\%), and when-question (8\%).

\noindent\textbf{Human Writer v.s. Machine Generator.}
As shown in Table~\ref{Tab:mg_vs_hw}, there are 41.7\% questions which are machine-generated questions. Since we still give crowd workers the freedom to modify questions if necessary, we cannot guarantee these questions are unchanged. We find that 33.56\% of our recommended questions have not been changed (100\% fuzzy matching score) and 19.92\% of them are slightly modified (81\%-99\% fuzzy matching score).
To dive into the characteristics and differences of these two question sources, we further conduct the human evaluation by sampling 200 conversation chunks randomly. We select chunks that have QG questions unchanged (i.e., sampling from the 33.56\% QG questions). We ask three annotators to first write an answer to the given question and conversation, then label fluency (how fluent and grammatically correct the question is, from 0 to 2), complexity (how hard to find an answer, from 0 to 2), and confidence (whether they are confident with their answer, 0 or 1). More details of each evaluation dimension (Section~\ref{sec:hwqg}) and performance difference (Table~\ref{tab:hwqg}) are shown in the Appendix. The results in Table~\ref{Tab:mg_vs_hw} indicate that QG questions are longer, more fluent, more complex, and crowd workers are less confident that they are providing the right answers. This observation further confirmed our hypothesis that the question generation strategy is effective to collect harder QA examples. 

\begin{table}[t]
\centering
\resizebox{\linewidth}{!}{
\begin{tabular}{lcccccc}
\hline
Source & \multicolumn{4}{c}{\textbf{Question Generator}} & \multicolumn{2}{c}{\textbf{Human Writer}} \\ \hline
Questions & \multicolumn{4}{c}{14,426 (41.7\%)} & \multicolumn{2}{c}{20,178 (58.3\%)} \\
Type & 100 & 81-99 & 51-79 & 0-50 & Ans. & Unans. \\
Ratio & 33.56\% & 19.92\% & 24.72\% & 21.80\% & 91.39\% & 8.61\% \\ \hline
Avg. Words & \multicolumn{4}{c}{\textbf{12.94} ($\pm 5.14$)} & \multicolumn{2}{c}{10.98 ($\pm 3.58$)} \\
Fluency & \multicolumn{4}{c}{\textbf{1.808}} & \multicolumn{2}{c}{1.658} \\
Complexity & \multicolumn{4}{c}{\textbf{0.899}} & \multicolumn{2}{c}{0.674} \\
Confidence & \multicolumn{4}{c}{\textbf{0.830}} & \multicolumn{2}{c}{0.902} \\ \hline
\end{tabular}
}
\caption{HW v.s. MG: Ratio and human evaluation.}
\label{Tab:mg_vs_hw}
\end{table}


\subsubsection{Answer Analysis} 
Following \citet{rajpurkar-etal-2016-squad}, we used Part-Of-Speech (POS)~\citep{kitaev-klein-2018-constituency} and Spacy NER taggers to study answers diversity. Firstly, we use the NER tagger to assign an entity type to the answers. However, since our answers are not necessary to be an entity, those answers without entity tags are then pass to the POS tagger, to extract the corresponding phrases tag.  In Table~\ref{Tab:diversityAnw}, we can see that Noun phrases make up 30.4\% of the data; followed by People, Organization, Dates, other numeric, and Countries; and the remaining are made up of clauses and other types. Full category distribution is shown in the Appendix (Figure~\ref{fig:diversity}). Note that there are around 1\% of answers in our dataset are coming from multiple source text spans (examples are shown in Appendix Table~\ref{Tab:multi-span-answers}). 

\subsection{Chunk Mode and Full Mode}
The main difference between the two modes is whether the conversational chunk we used to collect QA pairs is provided or not. In the chunk mode, our task is more like a traditional machine reading comprehension task that answers can be found (or cannot be found) in a short paragraph, usually less than 500 words. In the full mode, on the other hand, we usually need an information retrieval stage before the QA stage. For example, in the Natural Question dataset~\citep{kwiatkowski-etal-2019-natural}, they split Wikipedia into millions of passages and retrieve the most relevant one to answer. 

We define our full mode task with the following assumptions: 1) for the email and panel data, we assume to know which dialogue a question is corresponding to, that is, we only search chunks within the dialogue instead of all the possible conversations. This is simpler and more reasonable because each conversation is independent; 2) for slack data, we assume that we only know which channel a question belongs to but not the corresponding thread, so the retrieval part has to be done in the whole channel. Although chunk mode may be a better way to evaluate the ability of machine reading comprehension, the full mode is more practical as it is close to our setup in the real world.

\begin{table}[t]
\centering
\resizebox{\linewidth}{!}{
\begin{tabular}{lcl}
\hline
\multicolumn{1}{c}{\textbf{Answer type}} & \textbf{Percentage} & \multicolumn{1}{c}{\textbf{Example}} \\ \hline
Prepositional Phrase & 1.3\% & with `syntax-local-lift-module` \\
Nationalities or religious & 1.3\% & white Caucasian American \\
Monetary values & 1.6\% & \$250,000 \\
Clause & 5.4\% & need to use an external store for state \\
Countries, cities, states & 8.9\% & Chicago \\
Other Numeric & 9.6\% & page 66, volume 4 \\
Dates & 9.6\% & 2020 \\
Organizations & 11.4\% & Drug Enforcement Authority \\
People, including fictional & 12.5\% & Tommy Norment \\
Noun Phrase & 30.4\% & the Pulitzer Prize \\ \hline
\end{tabular}
}
\caption{Answer type analysis.}
\label{Tab:diversityAnw}
\end{table}

\section{Experimental Results}

\subsection{State-of-the-art Baselines}
There are two categories of question answering models: span-based extractive models which predict answers' start and end positions, and free-form text generation models which directly generate answers token by token. All the state-of-the-art models are based on large-scale language models, which are first pretrained on the general text and then finetuned on other QA tasks. We evaluate all of them on both zero-shot and finetuned settings (further finetuned on the {\dataset} training set), and both chunk mode and full mode with retrievers.
In addition, we run these models on the Molweni~\citep{li2020molweni} dataset for comparison and find out our baselines outperform the best-reported model, DADgraph~\citep{li2021dadgraph} model, which used expensive discourse annotation on graph neural network. We show the Molweni results in the Appendix (Table~\ref{Tab:results_molweni}).

\begin{table*}[t]
\centering
\resizebox{0.75\linewidth}{!}{
\begin{tabular}{l|ccc|ccc}
\hline
\textbf{} & \multicolumn{3}{c|}{Zero-Shot} & \multicolumn{3}{c}{Finetune} \\
 & \textbf{EM} & \textbf{F1} & \textbf{FZ-R} & \textbf{EM} & \textbf{F1} & \textbf{FZ-R} \\ \hline
Human Performance* & 79.99 & 89.87 & 92.33 & - & - & - \\ \hline
DistilBERT-Base-SQuAD2.0 & 40.04 & 46.90 & 59.62 & 57.28 & 68.88 & 75.39 \\
BERT-Base-SQuAD2.0 & 36.22 & 44.57 & 57.72 & 58.84 & 71.02 & 77.03 \\
BERT-Large-SQuAD2.0 & \textbf{53.54} & \textbf{62.58} & \textbf{71.11} & 64.93 & 76.65 & 81.27 \\
RoBERTa-Base-SQuAD2.0 & 48.92 & 57.33 & 67.40 & 63.64 & 75.53 & 80.38 \\
RoBERTa-Large-SQuAD2.0 & 50.78 & 59.73 & 69.11 & \textbf{67.80} & \textbf{78.80} & \textbf{83.10} \\ \hline
T5-Base-UnifiedQA & 51.95 & 65.48 & 73.26 & 64.98 & 76.52 & 81.69  \\
T5-Large-UnifiedQA & 58.81 & 71.67 & 77.72 & 66.76 & 78.67 & 83.21  \\
T5-3B-UnifiedQA &  \textbf{59.93} & \textbf{73.07} & \textbf{78.89} & \textbf{67.41} & \textbf{79.41} & \textbf{83.64}  \\ 
T5-11B-UnifiedQA & 44.96 & 61.52 & 68.68 & - & - & -  \\
\hline
\end{tabular}
}
\caption{Evaluation results: Chunk mode on the test set.}
\label{Tab:results_chunk_mode}
\end{table*}

\subsubsection{Span-based Models}
We use several models finetuned on the SQuAD 2.0 dataset as span extractive baselines. We use uploaded models from huggingface~\citep{wolf2019huggingface} library. DistilBERT~\citep{sanh2019distilbert} is a knowledge-distilled version with 40\% size reduction from the BERT model, and it is widely used in mobile devices. The BERT-Base and RoBERTa-Base~\citep{liu2019roberta} models are evaluated as the most commonly used in the research community. We also run the BERT-Large and RoBERTa-Large models as stronger baselines. We use the whole-word masking version of BERT-Large instead of the token masking one from the original paper since it performs better.

\subsubsection{Free-form Models}
We run several versions of UnifiedQA models~\citep{khashabi-etal-2020-unifiedqa} as strong generative QA baselines. UnifiedQA is based on T5 model~\citep{raffel2019exploring}, a language model that has been pretrained on 750GB C4 text corpus. UnifiedQA further finetuned T5 models on eight existing QA corpora spanning four diverse formats, including extractive, abstractive, multiple-choice, and yes/no questions. It has achieved state-of-the-art results on 10 factoid and commonsense QA datasets. We finetune UnifiedQA on our datasets with T5-Base, T5-Large size, and T5-3B. We report T5-11B size for the zero-shot performance. 

\begin{table*}[t]
\centering
\resizebox{0.75\linewidth}{!}{
\begin{tabular}{l|ccc|ccc}
\hline
\multicolumn{1}{c|}{BM25} & \multicolumn{3}{c|}{Zero-Shot} & \multicolumn{3}{c}{Finetune} \\
 & \textbf{EM} & \textbf{F1} & \textbf{FZ-R} & \textbf{EM} & \textbf{F1} & \textbf{FZ-R} \\ \hline
DistilBERT-Base-SQuAD2.0 & 29.36 & 34.09 & 50.35 & 39.39 & 48.38 & 60.46 \\ 
BERT-Base-SQuAD2.0 & 25.84 & 31.52 & 48.28 & 40.02 & 49.39 & 61.13 \\ 
BERT-Large-SQuAD2.0 & \textbf{37.09} & \textbf{43.44} & \textbf{57.21} & 44.50 & 53.48 & 64.21 \\ 
RoBERTa-Base-SQuAD2.0 & 34.61 & 40.74 & 55.37 & 43.18 & 52.64 & 63.62 \\ 
RoBERTa-Large-SQuAD2.0 & 35.54 & 41.50 & 55.79 & \textbf{45.59} & \textbf{54.42} & \textbf{65.23} \\  \hline
T5-Base-UnifiedQA &  36.47 & 47.11 & 59.22  & 43.95 & 52.96 & 64.22   \\  
T5-Large-UnifiedQA &  40.62 & 50.87 & 62.10  & 45.34 & 54.49 & 65.47 \\ 
T5-3B-UnifiedQA &  \textbf{41.76} & \textbf{52.68} & \textbf{63.54}  & \textbf{45.86} & \textbf{55.17} & \textbf{65.76} \\  \hline
\end{tabular}
}
\caption{Evaluation results: Full mode with BM25 retriever on the test set.}
\label{Tab:results_full_mode}
\end{table*}

\begin{table}[t]
\centering
\resizebox{0.9\linewidth}{!}{
\begin{tabular}{lcccc}
\hline
 & R@1 & R@3 & R@5 & R@10 \\ \hline
BM25 & 0.580 & 0.752 & 0.800 & 0.848 \\
DPR-wiki & 0.429 & 0.601 & 0.661 & 0.740 \\ \hline
\end{tabular}
}
\caption{BM25 and DPR-wiki result on the  test set.}
\label{Tab:results_retrieve}
\end{table}

\subsubsection{Retrieval Models}
Two retrieval baselines are investigated in this paper: BM25 and DPR-wiki~\citep{karpukhin-etal-2020-dense}. The BM25 retriever is a bag-of-words retrieval function weighted by term frequency and inverse document frequency. The DPR-wiki model is a BERT-based dense retriever model trained for open-domain QA tasks, learning to retrieve the most relevant Wikipedia passage.

\subsubsection{Computational Details}
We train most of our experiments on 2 V100 NVIDIA GPUs with a batch size that maximizes their memory usage, except T5-3B we train on four A100 NVIDIA GPUs with batch size 1 with several parallel tricks, such as fp16, sharded\_ddp and deepseep library. We train 10 epochs for all T5 models and 5 epochs for all BERT-based models. We release hyper-parameter setting and trained models to help reproduce baseline results.


\begin{table*}[t]
\centering
\resizebox{0.85\linewidth}{!}{
\begin{tabular}{l|cc|cc|cc|cc}
\hline
 & \multicolumn{4}{c|}{Zero-Shot} & \multicolumn{4}{c}{Finetune} \\ \cline{2-9} 
 & \multicolumn{2}{c|}{Ans.} & \multicolumn{2}{c|}{Unans. Binary} & \multicolumn{2}{c|}{Ans.} & \multicolumn{2}{c}{Unans. Binary} \\
 & \textbf{EM} & \textbf{F1} & \textbf{Recall} & \textbf{F1} & \textbf{EM} & \textbf{F1} & \textbf{Recall} & \textbf{F1} \\ \hline
DistilBERT-Base (SQuAD) & 38.12 & 45.32 & 77.97 & 16.84 &  57.81 & 70.00 & 46.89 & 40.85\\
BERT-Base (SQuAD2) & 34.07 & 42.84 & 78.53 &  16.17 & 59.18 & 71.98 & 51.98 & 43.36 \\
BERT-Large (SQuAD2) & \textbf{52.15} & \textbf{61.66} & 80.79   & \textbf{24.41}  & 65.44  & 77.76  & 54.80  & 49.39 \\
RoBERTa-Base (SQuAD2) & 47.50 & 56.34 &  76.84 & 20.28  & 64.32 & 76.81 & 50.28 & 46.19 \\
RoBERTa-Large (SQuAD2) & 48.91 & 58.32 & \textbf{87.57}  & 23.18  & \textbf{68.25} & \textbf{79.81} & \textbf{58.76} & \textbf{54.55} \\ \hline
T5-Base-UnifiedQA & 54.59 & 68.81 & 0.0  & 0.0 & 65.99 & 78.11 & 45.20 & 43.30 \\
T5-Large-UnifiedQA & 61.80 & 75.31 & 0.0 & 0.0  & 67.54 & 80.05 & 51.41 & 51.17 \\
T5-3B-UnifiedQA & \textbf{62.97} & \textbf{76.78} & 0.0 & 0.0  & \textbf{67.74} & \textbf{80.35} & \textbf{61.02} & \textbf{55.21} \\ \hline
\end{tabular}
}
\caption{Answerable/Unanswerable results: Chunk mode on the test set.}
\label{Tab:results_ans_unans}
\end{table*}

\subsection{Evaluation Metrics}
We follow the standard evaluation metrics in the QA community: exact match (EM) and F1 scores. The EM score is a strict score that predicted answers have to be the same as the ground truth answers. The F1 score is calculated by tokens overlapping between predicted answers and ground truth answers. In addition, we also report the FZ-R scores, which used the Levenshtein distance to calculate the differences between sequences. We follow \citet{rajpurkar-etal-2016-squad} to normalize the answers in several ways: remove stop-words, remove punctuation, and lowercase each character. We add one step with the \textit{num2words} and \textit{word2number} libraries to avoid prediction difference such as ``2'' and ``two''. 

\subsection{Performance Analysis}

\subsubsection{Chunk Mode}
As the chunk mode results on the test set shown in Table~\ref{Tab:results_chunk_mode}, UnifiedQA T5 models, in general, outperform BERT/RoBERTa models in the zero-shot setting, and the performance increases as the size of the model increases.
This observation matches the recent trend that large-scale pretrained language model finetuned on aggregated datasets of a specific downstream task (e.g., QA tasks~\citep{khashabi-etal-2020-unifiedqa} or dialogue task~\citep{wu-etal-2020-tod}) can show state-of-the-art performance by knowledge transfer.
Due to the space limit, all the development set results are shown in the Appendix.

We observe a big improvement from all the baselines after finetuning on our training set, suggesting the effectiveness of our data to improve dialogue understanding. Those span-based models, meanwhile, achieve similar performance to UnifiedQA T5 models with smaller model sizes.
BERT-Base model has the largest improvement gain by 22.6 EM score after finetuning.
We find that the UnifiedQA T5 model with 11B parameters cannot achieve performance as good as the 3B model, we guess that the released checkpoint has not been optimized well by~\citet{khashabi-etal-2020-unifiedqa}. 
In addition, we estimate human performance by asking crowd workers to answer around 10\% QA pairs in test set. We collect two answers for each question and select one that has a higher FZ-R score. We observe an EM score at around 80\% and an F1 score at 90\%, which still shows a considerable gap with existing models.


\subsubsection{Full Mode}
The retriever results are shown in Table~\ref{Tab:results_retrieve}, in which we find that BM25 outperforms DPR-wiki by a large margin in our dataset on the recall@$k$ measure, where we report $k=1, 3, 5, 10$. The two possible reasons are that 1) the difference in data distribution between Wikipedia and conversation is large and DPR is not able to properly transfer to unseen documents, and 2) questions in {\dataset} are more specific to those mentioned entities, which makes the BM25 method more reliable. We show the full mode results in Table~\ref{Tab:results_full_mode} using BM25 (DPR-wiki results in the Appendix Table~\ref{Tab:results_full_mode_dpr}). We use the top one retrieved conversational chunk as input to feed the trained QA models. As a result, the performance of UnifiedQA (T5-3B) drops by 18.2\% EM score in the zero-shot setting, and the finetuned results of RoBERTa-Large drop by 22.2\% EM score as well, suggesting a serious error propagation issue in the full mode that requires further investigation in the future work. 




\section{Error Analysis}
We further check the results difference between answerable and unanswerable questions in Table~\ref{Tab:results_ans_unans}. 
The UnifiedQA T5 models outperform span-based models among the answerable questions, however, they are not able to answer any unanswerable questions and keep predicting some ``answers''. More interestingly, we observe that those span-based models perform poorly on answerable questions, as they can achieve a high recall but a low F1 score on unanswerable questions with a binary setting (predict answerable or unanswerable). This implies that existing span-based models tend to predict our task as unanswerable, revealing their weakness of dialogue understanding ability. 

Then we check what kinds of QA samples in the test set are improved the most while finetuning on our training data using RoBERTa-Large. We find that 75\% of such samples are incorrectly predicted to be unanswerable, which is consistent with the results in Table~\ref{Tab:results_ans_unans}. We also analyze the error prediction after finetuning. We find that 35.5\% are what-question errors, 18.2\% are which-question errors, 12.1\% are how-question errors, and 10.3\% are who-question errors. 

In addition, we sample 100 QA pairs from the errors which have an FZ-R score lower than 50\% and manually check and categorize these predicted answers. We find out that 20\% of such examples are somehow reasonable and may be able to count as correct answers (e.g., UCLA v.s. University of California, Jay Sonneburg v.s. Jay), 31\% are predicted wrong answers but with correct entity type (e.g., Eurasia v.s. China, Susan Flynn v.s. Sara Shackleton), 38\% are wrong answers with different entity types (e.g., prison v.s. drug test, Thanksgiving v.s., fourth quarter), and 11\% are classified as unanswerable questions wrongly. This finding reveals the weakness of current evaluation metrics that they cannot measure semantic distances between two different answers.

\section{Related Work}
QA datasets can be categorized into four groups. The first one is cloze-style QA where a model has to fill in the blanks. For example, the Children’s Book Test~\citep{hill2015goldilocks} and the Who-did-What dataset~\citep{onishi-etal-2016-large}. The second one is reading comprehension QA where a model picks the answers for multiple-choice questions or a yes/no question. For examples, RACE~\citep{lai-etal-2017-race} and DREAM~\citep{sun-etal-2019-dream} datasets. The third one is span-based QA, such as SQuAD~\citep{rajpurkar-etal-2016-squad} and MS MARCO~\citep{nguyen2016ms} dataset, where a model extracts a text span from the given context as the answer. The fourth one is open-domain QA, where the answers are selected and extracted from a large pool of passages, e.g., the WikiQA~\citep{yang-etal-2015-wikiqa} and Natural Question~\citep{kwiatkowski-etal-2019-natural} datasets. 

Conversation-related QA tasks have focused on asking sequential questions and answers like a conversation and are grounded on a short passage. 
DoQA~\citep{campos2020doqa} is collected based on Stack Exchange, CoQA~\citep{reddy-etal-2019-coqa} and QuAC~\citep{choi-etal-2018-quac} are the two most representative conversational QA datasets under this category. 
CoQA contains conversational QA pairs, free-form answers along with text spans as rationales, and text passages from seven domains.  
QuAC collected data by a teacher-student setting on Wikipedia sections and it could be open-ended, unanswerable, or context-specific questions. 
Closest to our work, Dream~\citep{sun-etal-2019-dream} is a multiple-choice dialogue-based reading comprehension examination dataset, but the conversations are in daily chit-chat domains between two people.
FriendsQA~\citep{yang-choi-2019-friendsqa} is compiled from transcripts of the TV show Friends, which is also chit-chat conversations among characters and only has around one thousand dialogues.
Molweni~\citep{li2020molweni} is built on top of Ubuntu corpus~\citep{lowe2015ubuntu} for machine-reading comprehension tasks, but its conversations are short and focused on one single domain, and their questions are less diverse due to their data collection strategy (10 annotators).

In general, our task is also related to conversations as a knowledge source.
The dialogue state tracking task in task-oriented dialogue systems can be viewed as one specific branch of this goal as well, where tracking slots and values can be re-framed as a QA task~\citep{mccann2018natural, li2021zero}.
Moreover, extracting user attributes from open-domain conversations~\citep{wu2019getting}, getting to know the user through conversations, can be marked as one of the potential applications. The very recently proposed query-based meeting summarization dataset, QMSum~\citep{zhong2021qmsum}, can be viewed as one application of treating conversations as databases and conduct an abstractive question answering task.

\section{Conclusion}
{\dataset} is a new dataset that conducts QA on informative conversations such as emails, panels, and channels. We show the unique challenges of our tasks in both chunk mode with oracle partial conversations and full mode with a retrieval stage. We find that state-of-the-art QA models have limited dialogue understanding and tend to predict our answerable QA pairs as unanswerable. We provide a new testbed for QA on conversation tasks to facilitate future research. 

\section*{Ethical Considerations}
The {\dataset} benchmark proposed in this work could be helpful in creation of more powerful conversation retrieval and QA on conversations.
However, {\dataset} benchmark only covers a few domains as background conversations. Furthermore, even with our best efforts to ensure high quality and accuracy, the dataset might still contain incorrect labels and biases in some instances, which could be the inherent mistakes from the original dialogue datasets.
This could pose a risk if models that are evaluated or built using this benchmark are used in domains not covered by the dataset or if they leverage evidence from unreliable or biased dialogues.
Thus, the proposed benchmark should not be treated as a universal tool for all domains and scenarios.
We have used only the publicly available transcripts data and adhere to their guideline, for example, the Media data is for research-purpose only and cannot be used for commercial purpose. 
As conversations may have biased views, for example, specific political opinions from speakers, the transcripts and QA pairs will likely contain them. The content of the transcripts and summaries only reflect the views of the speakers, not the authors' point-of-views. 
We would like to remind our dataset users that there could have potential bias, toxicity, and subjective opinions in the selected conversations which may impact model training. 
Please view the content and data usage with discretion.


\bibliography{anthology,custom}
\bibliographystyle{acl_natbib}

\clearpage
\newpage

\appendix

\section{Appendix}
\label{sec:appendix}
\subsection{Dataset documentation and intended uses}

We follow datasheets for datasets guideline to document the followings.

\subsubsection{Motivation}
\begin{itemize}[leftmargin=*]
    \item For what purpose was the dataset created? Was there a specific task
in mind? Was there a specific gap that needed to be filled? 
    \begin{itemize}
        \item {\dataset} is created to test understanding of informative conversations such as business emails, panel discussions, and work channels. It is designed for QA on informative conversations to fill the gap of common Wikipedia-based QA tasks.
    \end{itemize}
    
    \item Who created the dataset (e.g., which team, research group) and on behalf of which entity (e.g., company, institution, organization)?
    \begin{itemize}
        \item Salesforce AI Research team and HKUST CAiRE team work together to create this dataset.
    \end{itemize}

    \item Who funded the creation of the dataset? If there is an associated grant, please provide the name of the grantor and the grant name and number.
    \begin{itemize}
        \item Salesforce AI research team funded the creation of the dataset.
    \end{itemize}


\end{itemize}

\subsubsection{Composition}
\begin{itemize}[leftmargin=*]
    \item What do the instances that comprise the dataset represent (e.g., documents, photos, people, countries)? Are there multiple types of instances (e.g., movies, users, and ratings; people and interactions between them; nodes and edges)? Please provide a description.

    \begin{itemize}
        \item {\dataset} has conversations (text) among speakers (people) and a set of corresponding QA pairs (text). 
    \end{itemize}

    \item How many instances are there in total (of each type, if appropriate)?
    \begin{itemize}
        \item {\dataset} has {\questionNumber} QA pairs and {\dialogNumber} conversations. Each conversation has 568.8 words in average and the longest one has 19,917 words.
    \end{itemize}
    
    \item Does the dataset contain all possible instances or is it a sample (not necessarily random) of instances from a larger set? If the dataset is a sample, then what is the larger set? Is the sample representative of the larger set (e.g., geographic coverage)? If so, please describe how this representativeness was validated/verified. If it is not representative of the larger set, please describe why not (e.g., to cover a more diverse range of instances, because instances were withheld or unavailable).
    \begin{itemize}
        \item The conversations in {\dataset} are randomly sampled from several conversational datasets, including BC3, Enron, Court, Media, and Slack, and the number of samples is decided based on related work and the budget.
    \end{itemize}

    \item What data does each instance consist of? “Raw” data (e.g., unprocessed text or images) or features? In either case, please provide a description.
    \begin{itemize}
        \item Each sample has raw text of conversations, speaker names, and QA pairs.
    \end{itemize}

    \item Is there a label or target associated with each instance? If so, please provide a description.
    \begin{itemize}
        \item Each answerable sample has at least one possible answer in a list format.
    \end{itemize}

    \item Is any information missing from individual instances? If so, please provide a description, explaining why this information is missing (e.g., because it was unavailable). This does not include intentionally removed information, but might include, e.g., redacted text.
    \begin{itemize}
        \item We do not include the crowd worker information due to the potential privacy issue.
    \end{itemize}

    \item Are relationships between individual instances made explicit (e.g., users’ movie ratings, social network links)? If so, please describe how these relationships are made explicit.
    \begin{itemize}
        \item N/A
    \end{itemize}

    \item Are there recommended data splits (e.g., training, development/validation, testing)? If so, please provide a description of these splits, explaining the rationale behind them.
    \begin{itemize}
        \item We provide official training, development, and testing splits.
    \end{itemize}

    \item Are there any errors, sources of noise, or redundancies in the
dataset? If so, please provide a description.
    \begin{itemize}
        \item There could have some potential noise of question or answer annotation.
    \end{itemize}

    \item Is the dataset self-contained, or does it link to or otherwise rely on external resources (e.g., websites, tweets, other datasets)? If it links to or relies on external resources, a) are there guarantees that they will exist, and remain constant, over time; b) are there official archival versions of the complete dataset (i.e., including the external resources as they existed at the time the dataset was created); c) are there any restrictions] (e.g., licenses, fees) associated with any of the external resources that might apply to a future user? Please provide descriptions of all external resources and any restrictions associated with them, as well as links or other access points, as appropriate.
    \begin{itemize}
        \item {\dataset} is self-contained.
    \end{itemize}

    \item Does the dataset contain data that might be considered confidential (e.g., data that is protected by legal privilege or by doctorpatient confidentiality, data that includes the content of individuals’ non-public communications)? If so, please provide a description.
    \begin{itemize}
        \item No, all the samples in {\dataset} is public available.
    \end{itemize}

    \item Does the dataset contain data that, if viewed directly, might be offensive, insulting, threatening, or might otherwise cause anxiety? If so, please describe why.
    \begin{itemize}
        \item No
    \end{itemize}

    \item Does the dataset relate to people? If not, you may skip the remaining questions in this section.
    \begin{itemize}
        \item Yes
    \end{itemize}

    \item Does the dataset identify any subpopulations (e.g., by age, gender)? If so, please describe how these subpopulations are identified and provide a description of their respective distributions within the dataset.
    \begin{itemize}
        \item {\dataset} contains different speakers with their names. Some samples have their role information, e.g., petitioner.
    \end{itemize}

    \item Is it possible to identify individuals (i.e., one or more natural persons), either directly or indirectly (i.e., in combination with other data) from the dataset? If so, please describe how.
    \begin{itemize}
        \item Yes, because some of the conversations are coming from public forums, therefore, people may be able to find the original speaker if they find the original media source.
    \end{itemize}

    \item Does the dataset contain data that might be considered sensitive in any way (e.g., data that reveals racial or ethnic origins, sexual. orientations, religious beliefs, political opinions or union memberships, or locations; financial or health data; biometric or genetic data; forms of government identification, such as social security numbers; criminal history)? If so, please provide a description.
    \begin{itemize}
        \item N/A. 
    \end{itemize}
    

\end{itemize}

\subsubsection{Collection Process}
\begin{itemize}[leftmargin=*]
    \item How was the data associated with each instance acquired? Was the data directly observable (e.g., raw text, movie ratings), reported by subjects (e.g., survey responses), or indirectly inferred/derived from other data (e.g., part-of-speech tags, model-based guesses for age or language)? If data was reported by subjects or indirectly inferred/derived from other data, was the data validated/verified? If so, please describe how.
    \begin{itemize}
        \item The QA data is collected by Amazon Mechanical Turk. The data is directly observable.
    \end{itemize}

    \item What mechanisms or procedures were used to collect the data (e.g., hardware apparatus or sensor, manual human curation, software program, software API)? How were these mechanisms or procedures validated? If the dataset is a sample from a larger set, what was the sampling strategy (e.g., deterministic, probabilistic with specific sampling probabilities)?
    \begin{itemize}
        \item The QA data is collected by Amazon Mechanical Turk, we design a user interface with instructions on the top and then given partial conversation as context.
    \end{itemize}

    \item Who was involved in the data collection process (e.g., students, crowdworkers, contractors) and how were they compensated (e.g., how much were crowdworkers paid)?
    \begin{itemize}
        \item Crowdworkers. We paid them roughly \$8-10 per hour, calculated by the average time to read and wriite one QA pair is approximately 4 minutes. 
    \end{itemize}

    \item Over what timeframe was the data collected? Does this timeframe match the creation timeframe of the data associated with the instances (e.g., recent crawl of old news articles)? If not, please describe the timeframe in which the data associated with the instances was created.
    \begin{itemize}
        \item The data was collected during Feb 2021 to March 2021.
    \end{itemize}

    \item Were any ethical review processes conducted (e.g., by an institutional review board)? If so, please provide a description of these review processes, including the outcomes, as well as a link or other access point to any supporting documentation.
    \begin{itemize}
        \item We have conduct an internal ethical review process by Salesforce ethical AI team, \url{https://einstein.ai/ethics}.
    \end{itemize}

    \item Does the dataset relate to people? If not, you may skip the remainder of the questions in this section.
    \begin{itemize}
        \item Yes.
    \end{itemize}

    \item Did you collect the data from the individuals in question directly, or obtain it via third parties or other sources (e.g., websites)?
    \begin{itemize}
        \item We obtain the data through AMT website.
    \end{itemize}

    \item Were the individuals in question notified about the data collection? If so, please describe (or show with screenshots or other information) how notice was provided, and provide a link or other access point to, or otherwise reproduce, the exact language of the notification itself.
    \begin{itemize}
        \item Yes, the turkers know the data collect procedure. Screenshots are shown Figure~\ref{fig:question_amthw}, Figure~\ref{fig:question_amtmg}, Figure~\ref{fig:question_amtv} in the Appendix.
    \end{itemize}

    \item Did the individuals in question consent to the collection and use of their data? If so, please describe (or show with screenshots or other information) how consent was requested and provided, and provide a link or other access point to, or otherwise reproduce, the exact language to which the individuals consented.
    \begin{itemize}
        \item AMT has its own data policy. \\         \url{https://www.mturk.com/acceptable-use-policy}.
    \end{itemize}

    \item If consent was obtained, were the consenting individuals provided with a mechanism to revoke their consent in the future or for certain uses? If so, please provide a description, as well as a link or other access point to the mechanism (if appropriate).
    \begin{itemize}
        \item \url{https://www.mturk.com/acceptable-use-policy}.
    \end{itemize}

    \item Has an analysis of the potential impact of the dataset and its use on data subjects (e.g., a data protection impact analysis) been conducted? If so, please provide a description of this analysis, including the outcomes, as well as a link or other access point to any supporting documentation.
    \begin{itemize}
        \item N/A
    \end{itemize}

\end{itemize}

\subsubsection{Preprocessing/cleaning/labeling}
\begin{itemize}[leftmargin=*]
    \item Was any preprocessing/cleaning/labeling of the data done (e.g., discretization or bucketing, tokenization, part-of-speech tagging, SIFT feature extraction, removal of instances, processing of missing values)? If so, please provide a description. If not, you may skip the. remainder of the questions in this section. 
    \begin{itemize}
        \item We conduct data cleaning such as removing code snippets before asking the crowd workers to provide corresponding QA pairs. Thus, no additional cleaning or preprocessing is done for the released dataset, only the reading scripts used to change the format for model reading are used.
    \end{itemize}
    
    \item Was the “raw” data saved in addition to the preprocessed/cleaned/labeled data (e.g., to support unanticipated future uses)? If so, please provide a link or other access point to the “raw” data.
    \begin{itemize}
        \item Yes, in the same link.
    \end{itemize}

    \item Is the software used to preprocess/clean/label the instances available? If so, please provide a link or other access point.
    \begin{itemize}
        \item \url{https://github.com/salesforce/QAConv}
    \end{itemize}

\end{itemize}

\subsubsection{Uses}
\begin{itemize}[leftmargin=*]
    \item Has the dataset been used for any tasks already? If so, please provide
a description.
    \begin{itemize}
        \item It is proposed to use for QA on conversations task.
    \end{itemize}

    \item Is there a repository that links to any or all papers or systems that use the dataset? If so, please provide a link or other access point.
    \begin{itemize}
        \item It is a new dataset. We run existing state-of-the-art models and release the code.
    \end{itemize}
    
    \item What (other) tasks could the dataset be used for?
    \begin{itemize}
        \item Many conversational AI related tasks can be applied or transferred, for examples, conversational retrieval and conversational machine reading.
    \end{itemize}
    
    \item Is there anything about the composition of the dataset or the way it was collected and preprocessed/cleaned/labeled that might impact future uses? For example, is there anything that a future user might need to know to avoid uses that could result in unfair treatment of individuals or groups (e.g., stereotyping, quality of service issues) or other undesirable harms (e.g., financial harms, legal risks) If so, please provide a description. Is there anything a future user could do to mitigate these undesirable harms?
    \begin{itemize}
        \item Different ways to disentangle conversations could impact the overall performance. In our current setting, we use and release the buffer-based chunking mechanism. 
    \end{itemize}
    
    \item Are there tasks for which the dataset should not be used? If so, please provide a description.
    \begin{itemize}
        \item Conversations from Media corpus should not be used for commercial usage.
    \end{itemize}
    
\end{itemize}

\subsubsection{Distribution}
\begin{itemize}[leftmargin=*]
    \item Will the dataset be distributed to third parties outside of the entity (e.g., company, institution, organization) on behalf of which the dataset was created? If so, please provide a description.
    \begin{itemize}
        \item No.
    \end{itemize}
    
    \item How will the dataset will be distributed (e.g., tarball on website, API, GitHub)? Does the dataset have a digital object identifier (DOI)?
    \begin{itemize}
        \item Release on Github. No DOI.
    \end{itemize}
    
    \item When will the dataset be distributed?
    \begin{itemize}
        \item Released.
    \end{itemize}
    
    \item Will the dataset be distributed under a copyright or other intellectual property (IP) license, and/or under applicable terms of use (ToU)? If so, please describe this license and/or ToU, and provide a link or other access point to, or otherwise reproduce, any relevant licensing terms or ToU, as well as any fees associated with these restrictions.
    \begin{itemize}
        \item BSD 3-Clause "New" or "Revised" License. 
    \end{itemize}
    
    \item Have any third parties imposed IP-based or other restrictions on the data associated with the instances? If so, please describe these restrictions, and provide a link or other access point to, or otherwise reproduce, any relevant licensing terms, as well as any fees associated with these restrictions.
    \begin{itemize}
        \item No.
    \end{itemize}
    
    \item Do any export controls or other regulatory restrictions apply to the dataset or to individual instances? If so, please describe these restrictions, and provide a link or other access point to, or otherwise reproduce, any supporting documentation.
    \begin{itemize}
        \item Media dataset is restricted their conversations to be research-only usage.
    \end{itemize}
    
\end{itemize}

\subsubsection{Maintenance}
\begin{itemize}[leftmargin=*]
    \item Who is supporting/hosting/maintaining the dataset?
    \begin{itemize}
        \item Salesforce AI Research team. Chien-Sheng (Jason) Wu is the corresponding author.
    \end{itemize}
    
    \item How can the owner/curator/manager of the dataset be contacted (e.g., email address)?
    \begin{itemize}
        \item Create an open issue on our Github repository or contact the authors.
    \end{itemize}
    
    \item Is there an erratum? If so, please provide a link or other access point.
    \begin{itemize}
        \item No.
    \end{itemize}
    
    \item Will the dataset be updated (e.g., to correct labeling errors, add new instances, delete instances)? If so, please describe how often, by whom, and how updates will be communicated to users (e.g., mailing list, GitHub)?
    \begin{itemize}
        \item No. If we plan to update in the future, we will indicate the information on our Github repository.
    \end{itemize}
    
    \item If the dataset relates to people, are there applicable limits on the retention of the data associated with the instances (e.g., were individuals in question told that their data would be retained for a fixed period of time and then deleted)? If so, please describe these limits and explain how they will be enforced.
    \begin{itemize}
        \item No.
    \end{itemize}
    
    \item Will older versions of the dataset continue to be supported/hosted/maintained? If so, please describe how. If not, please describe how its obsolescence will be communicated to users.
    \begin{itemize}
        \item Yes. If we plan to update the data, we will keep the original version available and then release the follow-up version, for example, {\dataset}-2.0
    \end{itemize}
    
    \item If others want to extend/augment/build on/contribute to the dataset, is there a mechanism for them to do so? If so, please provide a description. Will these contributions be validated/verified? If so, please describe how. If not, why not? Is there a process for communicating/distributing these contributions to other users? If so, please provide a description.
    \begin{itemize}
        \item Yes, they can submit a Github pull request or contact us privately. 
    \end{itemize}
    
\end{itemize}


  
  
  
  

\subsection{Test Data Additional Verification}
After random split, we run an additional verification step on the dev and test set. If the new collected answer is very similar with the original answer (FZR score > 90), we keep the original answer. If the new answer is similar within a margin (90 > FZR score > 75), we keep both answers. If the new answer is very different from the original answer (75 > FZR score), we will run one more verification step to get the 3rd answers. We pick the most similar two answers as the gold answers if their FZR score is > 75, otherwise, we manually looked into those controversial QA pairs and made the final judgement.


\subsection{License and Privacy}
\label{sec:license}
\begin{itemize}
 \item BC3: Creative Commons Attribution-Share Alike 3.0 Unported License.
 \item Enron: Creative Commons Attribution 3.0 United States license.
 \item Court: This material is based upon work supported in part by the National Science Foundation under grant IIS-0910664.  Any opinions, findings, and conclusions or recommendations expressed above are those of the author(s) and do not necessarily reflect the views of the National Science Foundation.
 
 \item Media: Only the publicly available transcripts data from the media sources are included. 
 
 \item Slack: Numerous public Slack chat channels (\url{https://slack.com/}) have recently become available that are focused on specific software engineering-related discussion topics.
 
\end{itemize}

\subsection{Human evaluation description of human-written and machine-generated questions.}
\label{sec:hwqg}
Rate [Fluency of the question]: 
\begin{itemize}
    \item (A) The question is fluent and has good grammar. I can understand clearly.
    \item (B) The question is somewhat fluent with some minor grammar errors. But it does not influence my reading.
    \item (C) The question is not fluent and has serious grammar error. I can hardly understand it.
\end{itemize}

Rate [Complexity of the question]: 
\begin{itemize}
    \item (A) The answer to the question is hard to find. I have to read the whole conversation back-and-forth more than one time.
    \item (B) The answer to the question is not that hard to find. I can find the answer by reading several sentences once.
    \item (C) The answer to the question is easy to find. I can find the answer by only reading only one sentence.
\end{itemize}

Rate [Confidence of the answer]: 
\begin{itemize}
    \item (A) I am confident that my answer is correct.
    \item (B) I am not confident that my answer is correct.
\end{itemize}

\begin{table}[h]
\centering
\resizebox{0.9\linewidth}{!}{
\begin{tabular}{lcccc}
\hline
 & R@1 & R@3 & R@5 & R@10 \\ \hline
BM25 & 0.586 & 0.757 & 0.802 & 0.852 \\
DPR-wiki & 0.424 & 0.590 & 0.660 & 0.741 \\ 
\hline
\end{tabular}
}
\caption{Retriever results: BM25 on the dev set.}
\end{table}

\begin{table*}[t]
\centering
\resizebox{0.7\linewidth}{!}{
\begin{tabular}{l|ccc|ccc}
\hline
 & \multicolumn{3}{c|}{Zero-Shot} & \multicolumn{3}{c}{Finetune} \\
 & \textbf{EM} & \textbf{F1} & \textbf{FZ-R} & \textbf{EM} & \textbf{F1} & \textbf{FZ-R} \\ \hline
Human Performance & 64.3 & 80.2 & - & - & - & - \\ \hline
DialogueGCN* & - & - & - & 45.7 & 61.0 & - \\
DADgraph* & - & - & - & 46.5 & 61.5 & - \\
BERT-Large-SQuAD2.0 & 3626 & 45.90 & 56.90 & 53.43 & 66.85 & 73.50 \\
RoBERTa-Large-SQuAD2.0 & \textbf{38.42} & 51.37 & 60.33 & \textbf{53.92} & 67.47 & 73.62 \\ \hline
T5-Large-UnifiedQA & 34.52 & \textbf{53.64} & 63.08 & 52.14 & 69.04 & \textbf{75.38} \\
T5-3B-UnifiedQA & 35.01 & 55.51 & \textbf{64.14} & 52.14 & \textbf{69.21} & 75.25 \\ \hline
\end{tabular}
}
\caption{Evaluation results: Molweni on the test set. * number is obtained from the original paper.}
\label{Tab:results_molweni}
\end{table*}

\begin{table*}[t]
\centering
\resizebox{0.8\linewidth}{!}{
\begin{tabular}{ll|ccc|ccc}
\hline
 & \textbf{} & \multicolumn{3}{c|}{Zero-Shot} & \multicolumn{3}{c}{Finetune} \\
 &  & EM & F1 & FZ-R & EM & F1 & FZ-R \\ \hline
QG & T5-Base-UnifiedQA & 45.63 & 58.27 & 67.90 & 61.20 & 72.04 & 77.99 \\
 & T5-Large-UnifiedQA  & 53.68 & 64.99 & 72.78 & 62.64 & 73.31 & 79.00 \\
 & T5-3B-UnifiedQA     & 55.81 & 66.85 & 74.30 & 62.41 & 73.35 & 78.80 \\
 \hline
HW & T5-Base-UnifiedQA& 55.50 & 69.53 & 76.27& 67.11 & 79.04 & 83.77 \\
 & T5-Large-UnifiedQA & 61.69 & 75.42 & 80.49 & 69.07 & 81.68 & 85.57 \\
 & T5-3B-UnifiedQA    & 62.24 & 76.56 & 81.46 & 70.22 & 82.82 & 86.36 \\
 \hline
\end{tabular}
}
\caption{QG v.s. HW questions: test set results}
\label{tab:hwqg}
\end{table*}

\begin{table*}[h]
\centering
\resizebox{0.8\linewidth}{!}{
\begin{tabular}{l|ccc|ccc}
\hline
\multicolumn{1}{c|}{DPR-wiki} & \multicolumn{3}{c|}{Zero-Shot} & \multicolumn{3}{c}{Fine-Tune} \\
 & \textbf{EM} & \textbf{F1} & \textbf{FZ-R} & \textbf{EM} & \textbf{F1} & \textbf{FZ-R} \\ \hline
DistilBERT-Base-SQuAD2.0 & 10.90 & 12.56 & 34.63 & 11.83 & 15.47 & 36.33 \\
BERT-Base-SQuAD2.0 & 9.48 & 11.03 & 33.49 & 11.75 & 15.64 & 36.71 \\
BERT-Large-SQuAD2.0 & 12.35 & 14.15 & 35.63 & 12.97 & 16.79 & 37.61 \\
RoBERTa-Base-SQuAD2.0 & 11.66 & 13.43 & 35.30 & 12.24 & 16.05 & 37.01 \\
RoBERTa-Large-SQuAD2.0 & 11.88 & 13.62 & 35.37 & 13.22 & 17.00 & 37.94\\ \hline
T5-Base-UnifiedQA & 8.93 & 14.65 & 35.31 & 12.70 & 16.70 & 37.64 \\
T5-Large-UnifiedQA& 10.30 & 16.10 & 36.46 & 13.41 & 17.50 & 38.14 \\
T5-3B-UnifiedQA   & 10.65 & 17.46 & 38.25 & 13.36 & 17.84 & 38.68 \\ 
\hline
\end{tabular}
}
\caption{Evaluation results: Full mode with DPR-wiki on the test set.}
\label{Tab:results_full_mode_dpr}
\end{table*}

\begin{table*}[t]
\centering
\resizebox{0.8\linewidth}{!}{
\begin{tabular}{l|ccc|ccc}
\hline
 & \multicolumn{3}{c|}{Zero-Shot} & \multicolumn{3}{c}{Finetune} \\
 & EM & F1 & FZ-R & EM & F1 & FZ-R \\ \hline
DistilBERT-Base-SQuAD2.0 &  39.92 & 47.66 & 60.50  &  56.72 & 69.26 & 76.06  \\
BERT-Base-SQuAD2.0 &  36.37 & 44.74 & 58.20  &  59.56 & 71.04 & 77.64 \\
BERT-Large-SQuAD2.0 & 52.27 & 61.46 & 70.37 & 64.21 & 75.95 & 81.25 \\
RoBERTa-Base-SQuAD2.0 & 50.25 & 59.25 & 68.95 & 63.03 & 74.93 & 80.47 \\
RoBERTa-Large-SQuAD2.0 & 51.26 & 60.78 & 70.02 & 66.17 & 77.87 & 83.00 \\ \hline
T5-Base-UnifiedQA & 51.45 & 65.99 & 73.47 & 63.77 & 76.22 & 81.28 \\
T5-Large-UnifiedQA& 58.20 & 71.45 & 77.85 & 66.07 & 78.53 & 83.33 \\
T5-3B-UnifiedQA   & 59.78 & 72.76 & 78.80 & 67.32 & 79.32 & 83.82 \\ 
T5-11B-UnifiedQA  & 45.14 & 61.55 & 69.12 & - & - & - \\ 
\hline
\end{tabular}
}
\caption{Evaluation results: Chunk mode on the dev set.}
\end{table*}

\begin{table*}[t]
\centering
\resizebox{0.8\linewidth}{!}{
\begin{tabular}{l|ccc|ccc}
\hline
 & \multicolumn{3}{c|}{Zero-Shot} & \multicolumn{3}{c}{Finetune} \\
 & EM & F1 & FZ-R & EM & F1 & FZ-R \\ \hline
DistilBERT-Base-SQuAD2.0  & 28.93 & 34.55 & 51.03 & 38.66 & 48.70 & 60.80 \\
BERT-Base-SQuAD2.0  & 26.20 & 32.22 & 49.14 & 40.25 & 49.58 & 61.72 \\
BERT-Large-SQuAD2.0  & 36.20 & 42.94 & 56.98 & 43.09 & 52.70 & 64.02 \\
RoBERTa-Base-SQuAD2.0  & 35.93 & 42.32 & 56.59 & 43.03 & 52.43 & 63.69 \\
RoBERTa-Large-SQuAD2.0  & 35.93 & 42.71 & 56.85 & 45.19 & 54.33 & 65.45 \\ \hline
T5-Base-UnifiedQA & 35.44 & 47.05 & 59.56 & 43.74 & 53.54 & 64.45 \\
T5-Large-UnifiedQA& 39.56 & 50.82 & 62.40 & 44.40 & 54.58 & 65.31 \\
T5-3B-UnifiedQA   & 40.79 & 52.11 & 63.63 & 46.37 & 56.16 & 66.59 \\ \hline
\end{tabular}
}
\caption{Evaluation results: Full mode with BM25 on the dev set.}
\end{table*}

\begin{table*}[h]
\centering
\resizebox{0.8\linewidth}{!}{
\begin{tabular}{l|ccc|ccc}
\hline
\multicolumn{1}{c|}{DPR-wiki} & \multicolumn{3}{c|}{Zero-Shot} & \multicolumn{3}{c}{Fine-Tune} \\
 & \textbf{EM} & \textbf{F1} & \textbf{FZ-R} & \textbf{EM} & \textbf{F1} & \textbf{FZ-R} \\ \hline
DistilBERT-Base-SQuAD2.0 & 11.04 & 12.32 & 34.83 & 11.64 & 15.23 & 36.61 \\
BERT-Base-SQuAD2.0 & 9.73 & 10.94 & 33.89 & 12.32 & 15.54 & 36.66 \\
BERT-Large-SQuAD2.0 & 13.01 & 14.41 & 36.35 & 13.31 & 16.69 & 37.62 \\
RoBERTa-Base-SQuAD2.0 & 12.40 & 13.76 & 35.93 & 13.11 & 16.46 & 37.47 \\
RoBERTa-Large-SQuAD2.0 & 12.57 & 13.97 & 35.92 & 13.77 & 16.90 & 37.89\\ \hline
T5-Base-UnifiedQA & 8.85 & 13.88 & 35.13 & 12.62 & 16.26 & 37.54 \\
T5-Large-UnifiedQA& 9.95 & 15.28 & 36.55 & 13.31 & 17.27 & 38.22 \\
T5-3B-UnifiedQA   & 11.04 & 16.97 & 38.16 & 14.04 & 17.74 & 38.72 \\ \hline
\end{tabular}
}
\caption{Evaluation results: Full mode with DPR-wiki on the dev set.}
\label{Tab:results_full_mode_dpr}
\end{table*}


\begin{table*}[t]
\centering
\resizebox{0.9\linewidth}{!}{
\begin{tabular}{lll}
\hline
\multicolumn{1}{c}{\textbf{Relevant Context}} & \multicolumn{1}{c}{\textbf{Question}} & \multicolumn{1}{c}{\textbf{Answer}} \\ \hline
\begin{tabular}[c]{@{}l@{}}... David Klinger: There's a term of art \\ called awful, but lawful. So sometimes\\  officers are involved in shootings that \\ don't really sound that good, but the law\\  says it was an appropriate ...\end{tabular} & what can be awful but lawful? & officer involved shootings \\ \hline
\begin{tabular}[c]{@{}l@{}}... one foreign government should not \\ be able to come into our courts and \\ enforce its sovereign power by using \\ our courts to collect taxes from our \\ citizens...\end{tabular} & \begin{tabular}[c]{@{}l@{}}how do one foreign government \\ should not be able to come into the \\ courts and enforce its sovereign power?\end{tabular} & \begin{tabular}[c]{@{}l@{}}by using the courts to \\ collect taxes from the citizens.\end{tabular} \\ \hline
\begin{tabular}[c]{@{}l@{}}... directly in your mutable set without \\ worrying about it, since there can only\\  be expansion in one module per visit \\ to your module. so you'll never end up\\  with `'module` being returned for two\\  different modules before your mutable\\  set is emptied. gonzalo: so, to ...\end{tabular} & \begin{tabular}[c]{@{}l@{}}how many expansions can be \\ in one module per visit?\end{tabular} & one expansion per visit \\ \hline
\end{tabular}
}
\caption{Examples of multi-span answers in {\dataset}}
\label{Tab:multi-span-answers}
\end{table*}

\begin{table*}[h]
\centering
\resizebox{0.9\linewidth}{!}{
\begin{tabular}{ll}
\hline
Partial Context & \begin{tabular}[c]{@{}l@{}}...\\ \\ Steve Duffy: ..., but I don't know if Enron would even consider this.  Studdert might have\textbackslash{}nthe best feel for this.  \\ Separately, the defendant group will get back to us\textbackslash{}non any offer they might be willing to make to settle just the \\ Montana case,\textbackslash{}nbut it appears that their real interest would be in a \textbackslash{}"global\textbackslash{}" deal.  Any\textbackslash{}ncomments?  SWD\\ \\ Michael Burke: Steve, Stan and I have discussed this and we agree that Mike Moran should\textbackslash{}ntake the lead and \\ explore all aspects of an Enron Global deal.  I know that\textbackslash{}nyou will assist Mike in this endeavor.  thanks, mike\\ \\ Steve Duffy: Sounds good.  Mike Moran has the numbers for our Montana lawyers and I will\textbackslash{}nassist him any \\ way I can.  The big question is whether Enron, as a whole,\textbackslash{}nwould be willing to give up any protection they might\\  still have under the\textbackslash{}nold  InterNorth policies.  SWD\\ \\ ...\end{tabular} \\
Question & What person has the numbers for the Montana lawyers and is best qualified to explore the deal? \\ \hline
Partial Context & \begin{tabular}[c]{@{}l@{}}...\\ \\ OFEIBEA QUIST-ARCTON, BYLINE: One woman we spoke to has lived here all her life. She was born here, \\ married here, has children here. She said I'm going. I don't feel safe. You know, the ground was shaking when we \\ heard those bombs. We don't feel ...\\ \\ JENNIFER LUDDEN, HOST: \\ We are talking about the tensions and violence in Nigeria. We'll have more with NPR's Ofeibea Quist-Arcton from \\ Nigeria, and also former Ambassador John Campbell coming up. We'll also talk with an activist from Nigeria. If \\ you have questions, ...\\ \\ JENNIFER LUDDEN, HOST: This is TALK OF THE NATION from NPR News. I'm Jennifer Ludden. Nigeria has \\ long faced challenges from corruption, an economy that relies on oil exports and simmering ethnic and religious tensions, \\ tensions made evident in the recent series of bombings by Boko Haram, the militant ...\\ \\ JENNIFER LUDDEN, HOST: \\ It's the latest crisis for President Goodluck Jonathan. We're talking today with Ofeibea Quist-Arcton, NPR's foreign \\ correspondent, now in Kano, Nigeria; and John Campbell, former U.S. ambassador and political counselor to Nigeria. \\ He's now a senior fellow for Africa policy studies at the Council on Foreign Relations.\\ ...\end{tabular} \\
Question & Who is the president of the country where Ofeibea quist-arcton is talking about the tensions and violence in Nigeria ? \\ \hline
Partial Context & \begin{tabular}[c]{@{}l@{}}...\\ \\ \\ Karoline: are you using pytest?  there are a couple of plugins for parallelization\\ Valeri: Yes pytest\\ Eliana: pytest-xdist is pretty good\\ Valeri: What does that do?\\ Karoline\\ : yeah that and \\ pytest-parallel are worth a look\\ .  basically they \\ allow you to paralelize your tests\\ Valeri: Okay\\ Valeri: Will definitely look into those\\ Valeri: Thanks \textless{}@Eliana\textgreater \textless{}@Karoline\textgreater ,taco,\\ \\ \\ …\end{tabular} \\
Question & What program allows the user to parallelize the tests and is recommended by Karoline? \\ \hline
Partial Context & \begin{tabular}[c]{@{}l@{}}…\\ \\ \\ MR. FREEDMAN (RESPONDENT): … They both deserve the death penalty. They -- they were -- the prosecutors \\ were aware that the -- the death penalty is what stirs the pot here, and so they were urging somebody to be the shooter \\ to get the death penalty. If this wasn't a death penalty case, I don't think they -- it would have mattered who killed who. \\ And so they were urging --\\ \\ JUSTICE KENNEDY: Well, I think there's quite a difference in -- in case A where you say our position is that Stumpf \\ was the shooter, pure and simple. That's it. In case B, they say we think Stumpf was the shooter. We're not 100 percent \\ sure, but he should get the death penalty. The alternative is before the sentencer and the sentencer can make that \\ determination.\\ \\ …\end{tabular} \\
Question & Which person was mentioned as the shooter in case A and B? \\ \hline
\end{tabular}
}

\caption{Examples of multi-hop questions}
\label{Tab:multi-hop-questions}
\end{table*}

\begin{figure*}[h]
    \centering
    \includegraphics[width=0.75\linewidth]{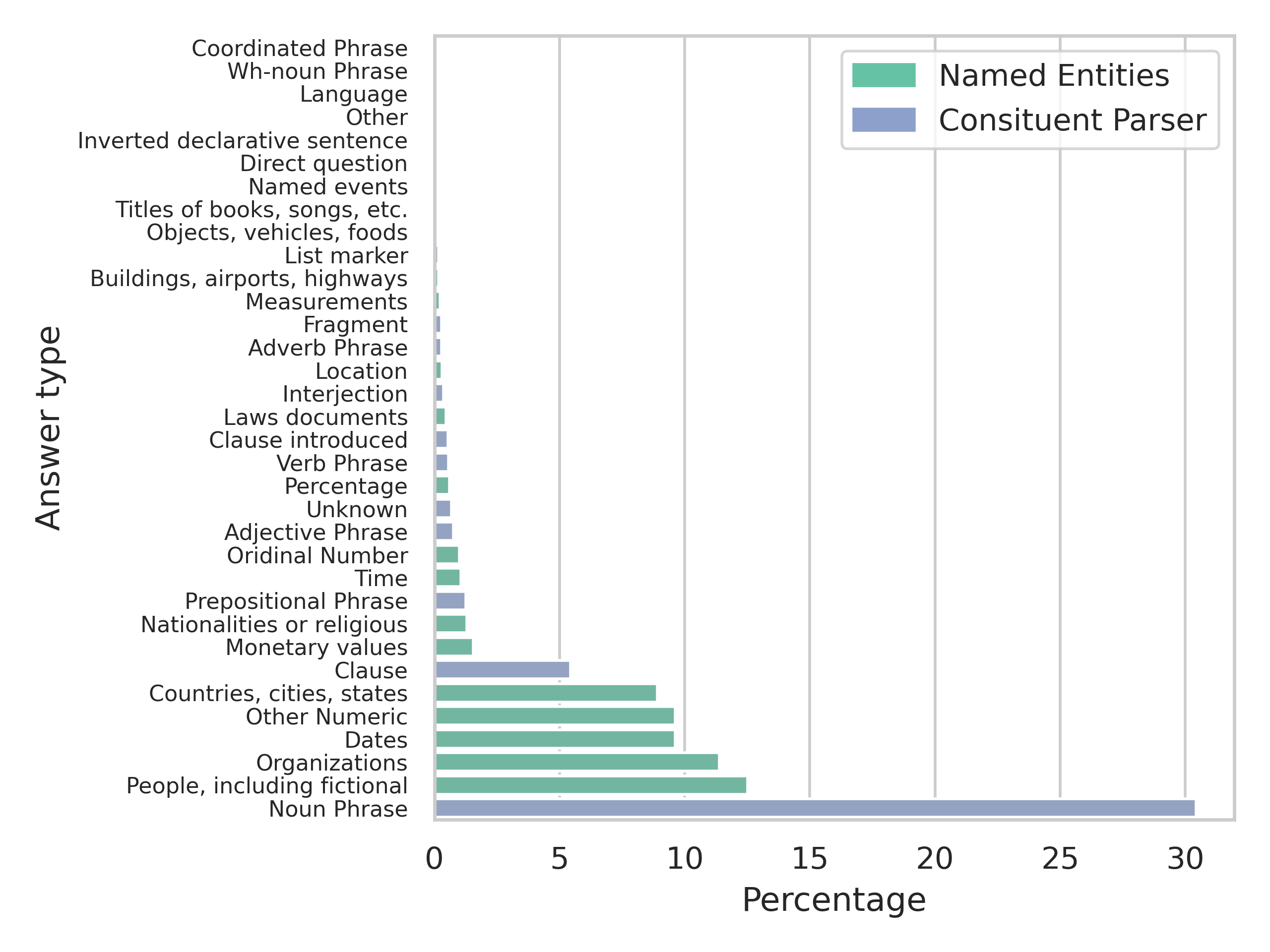}
    \caption{Diversity in answers in all categories.}
    \label{fig:diversity}
\end{figure*}

\begin{figure*}[t]
    \centering
    \includegraphics[width=0.9\linewidth]{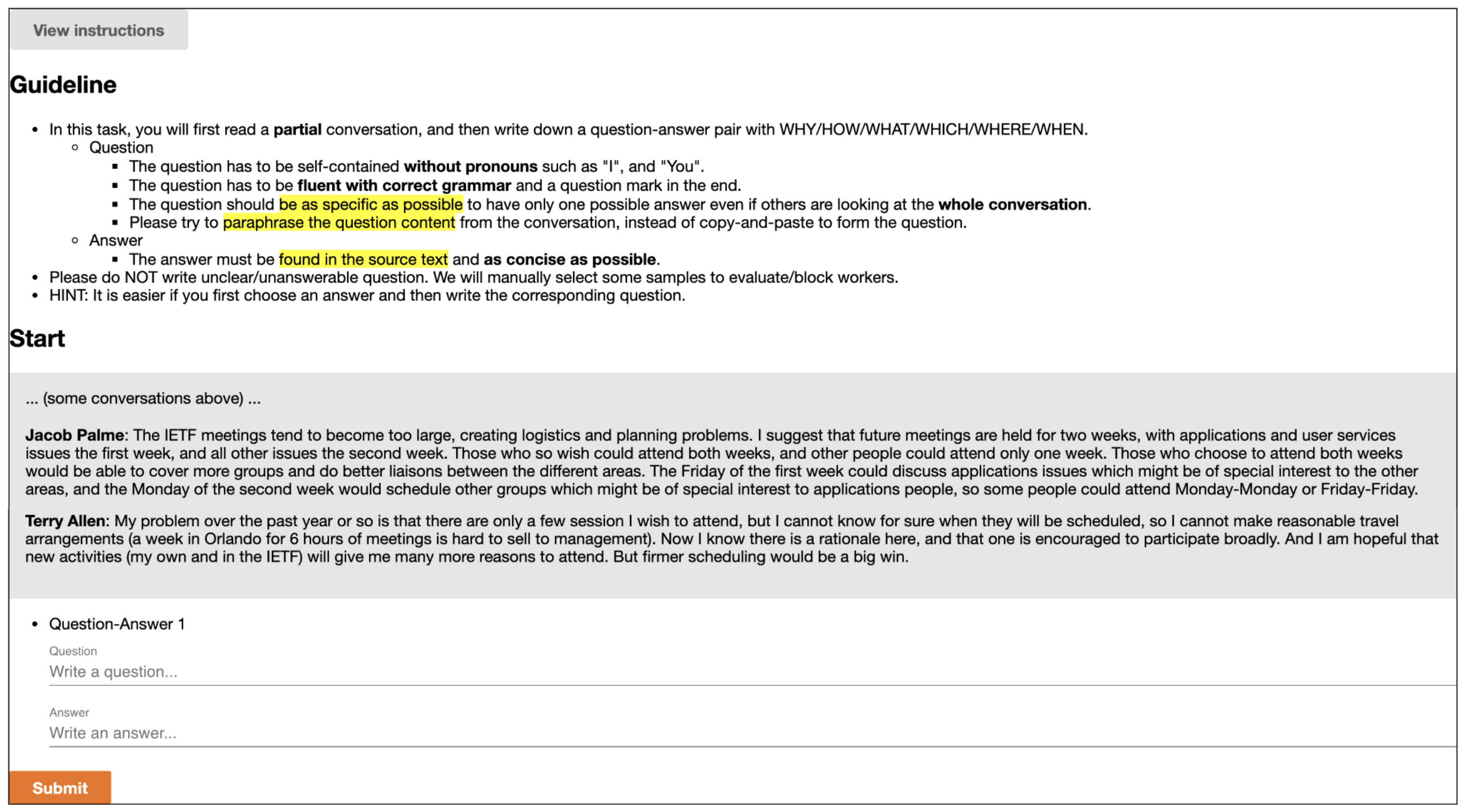}
    \caption{Screenshot for human-written QA collection.}
    \label{fig:question_amthw}
\end{figure*}

\begin{figure*}[h]
    \centering
    \includegraphics[width=0.9\linewidth]{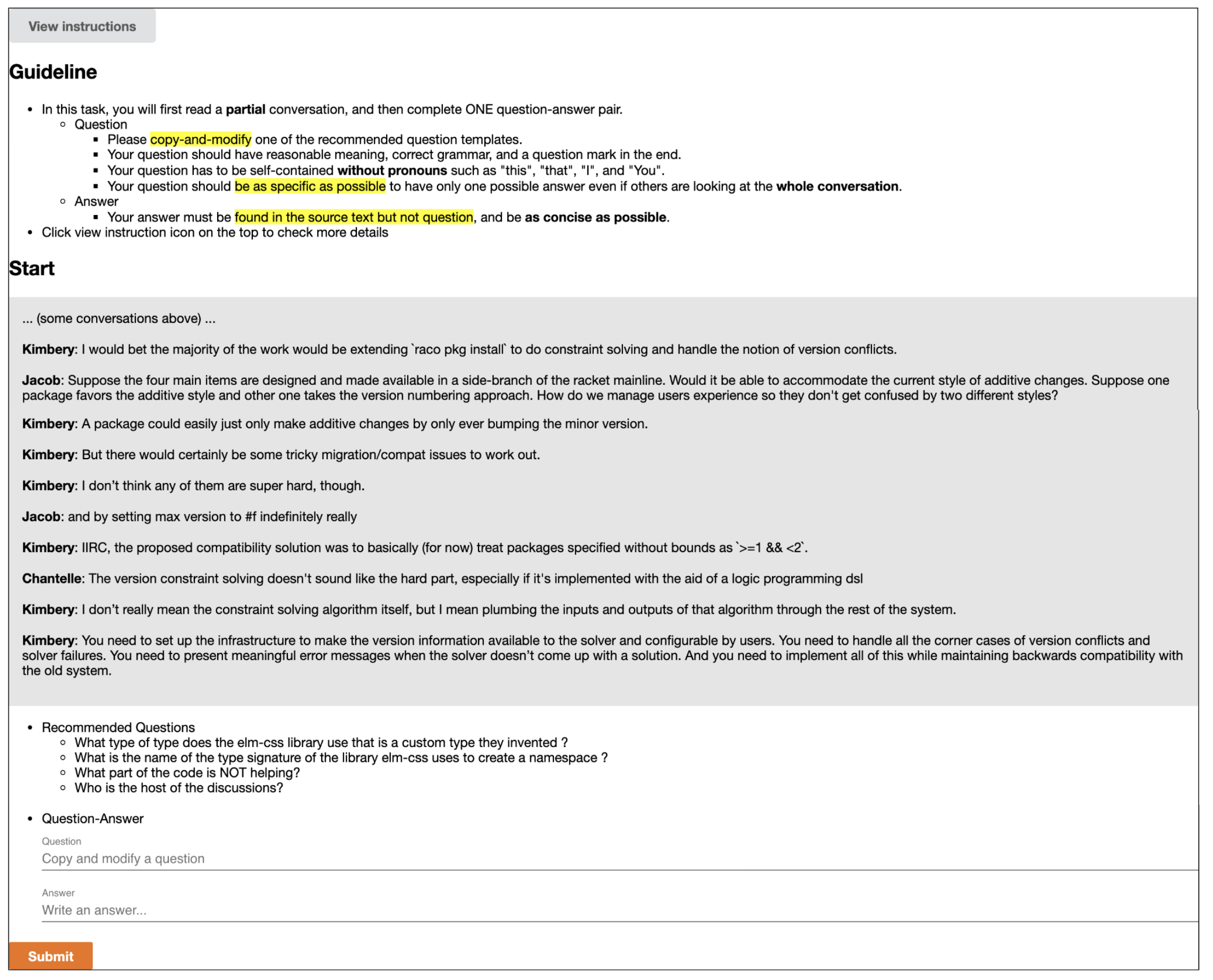}
    \caption{Screenshot for machine-generated QA collection.}
    \label{fig:question_amtmg}
\end{figure*}

\begin{figure*}[h]
    \centering
    \includegraphics[width=0.9\linewidth]{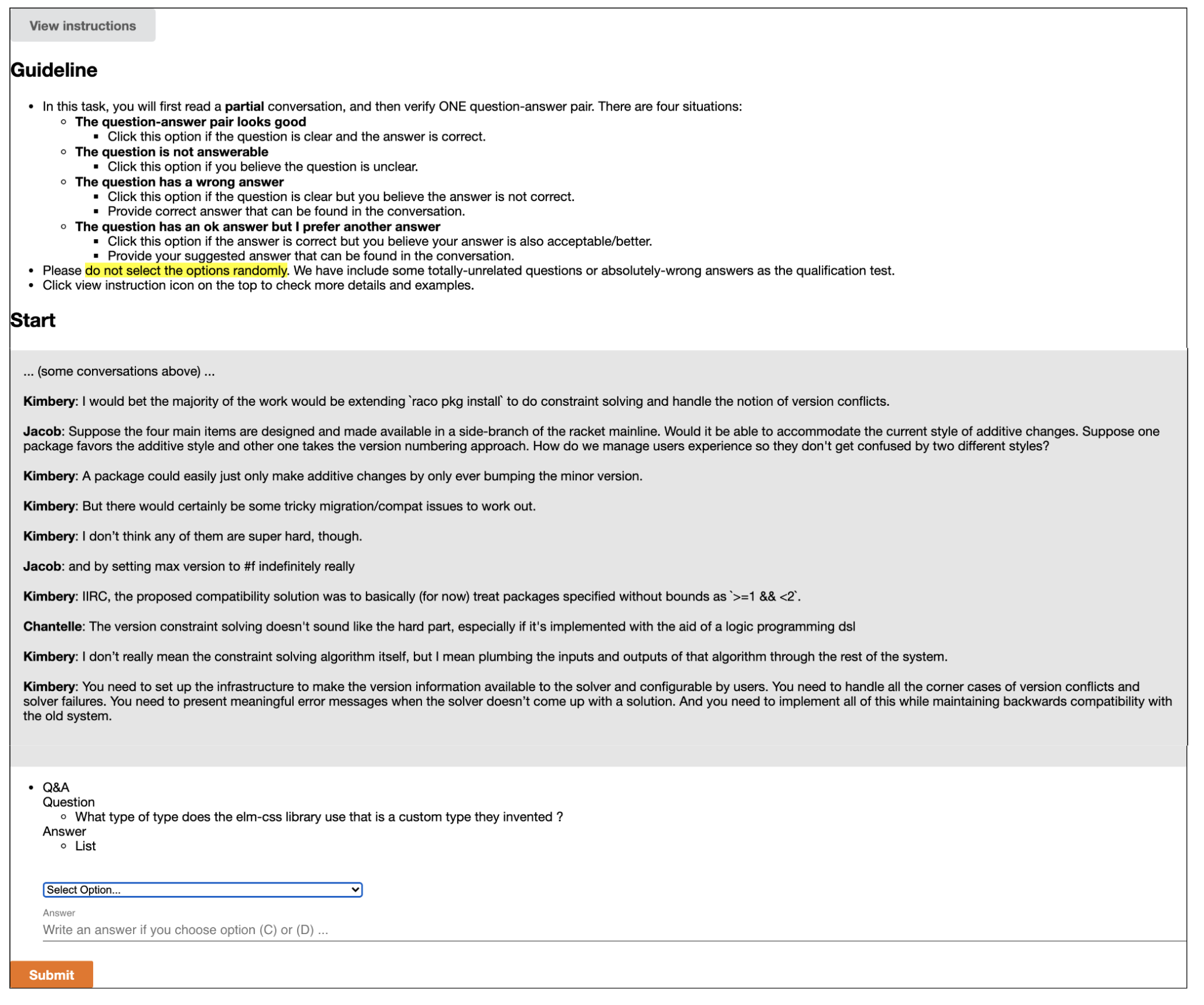}
    \caption{Screenshot for QA verification.}
    \label{fig:question_amtv}
\end{figure*}

\end{document}